\documentclass[conference]{IEEEtran}
\IEEEoverridecommandlockouts
\usepackage{cite}
\usepackage{amsmath,amssymb,amsfonts}
\usepackage{marvosym}
\usepackage{algorithmic}
\usepackage{graphicx}
\usepackage{textcomp}
\usepackage{xcolor}
\def\BibTeX{{\rm B\kern-.05em{\sc i\kern-.025em b}\kern-.08em
		T\kern-.1667em\lower.7ex\hbox{E}\kern-.125emX}}
\usepackage[utf8]{inputenc}
\usepackage[english]{babel}

\usepackage{enumerate}% http://ctan.org/pkg/enumerate
\usepackage{tikz,xcolor,hyperref}
\usepackage{marvosym}

\usepackage[utf8]{inputenc}
\usepackage[english]{babel}
\usetikzlibrary{positioning,chains,fit,shapes,calc}
\usetikzlibrary{
	graphs,
	graphs.standard
}

\hypersetup{draft}

\definecolor{lime}{HTML}{A6CE39}
\DeclareRobustCommand{\orcidicon}{%
	\begin{tikzpicture}
	\draw[lime, fill=lime] (0,0) 
	circle [radius=0.16] 
	node[white] {{\fontfamily{qag}\selectfont \tiny ID}};
	\draw[white, fill=white] (-0.0625,0.095) 
	circle [radius=0.007];
	\end{tikzpicture}
	\hspace{-2mm}
}

\foreach \x in {A, ..., Z}{%
	\expandafter\xdef\csname orcid\x\endcsname{\noexpand\href{https://orcid.org/\csname orcidauthor\x\endcsname}{\noexpand\orcidicon}}
}

\begin{document}
	
	\title{An Assignment Problem Formulation for Dominance Move Indicator  \\
		{\footnotesize \textsuperscript{*}}
		\thanks{$^1$ PPMMC - Programa de pós-graduação em modelagem matemática e computacional}
	}
	
	\author{\IEEEauthorblockN{1\textsuperscript{st}Claudio Lucio do Val Lopes}
		\IEEEauthorblockA{\textit{PPGMMC$^1$} \\
			\textit{CEFET-MG}\\
			BH, Brazil\\
			claudiolucio@gmail.com  \orcidA{} }
		\and
		\IEEEauthorblockN{2\textsuperscript{nd} Flávio Vinícius Cruzeiro Martins}
		\IEEEauthorblockA{\textit{PPGMMC$^1$} \\
			\textit{CEFET-MG}\\
			BH, Brazil \\
			flaviocruzeiro@cefetmg.br \orcidB{} }
		\and
		\IEEEauthorblockN{3\textsuperscript{rd} Elizabeth F. Wanner}
		\IEEEauthorblockA{\textit{PPGMMC$^1$ - CEFET-MG} \\
			{Computer Science Group, Aston University,} \\
			{Birmingham, UK} \\
			efwanner@decom.cefetmg.br \orcidC{} }
	}

	\maketitle
	
	\begin{abstract}
        Dominance move (DoM) is a binary quality indicator to compare solution sets in multiobjective optimization. The indicator allows a more natural and intuitive relation when comparing solution sets. Like the $\epsilon$-indicators, it is Pareto compliant and does not demand any parameters or reference sets. In spite of its advantages, the combinatorial calculation nature is a limitation. The original formulation presents an efficient method to calculate it in a bi-objective case only. This work presents an assignment formulation to calculate DoM in problems with three objectives or more. Some initial experiments, in the bi-objective space, were done to show that DoM has a similar interpretation as $\epsilon$-indicators, and to show that our model formulation is correct. Next, other experiments, using three dimensions, were also done to show how DoM could be compared with other indicators: inverted generational distance (IGD) and hypervolume (HV). The assignment formulation for DoM is valid not only for three objectives but for more. Finally, there are some strengths and weaknesses, which are discussed and detailed. 
	\end{abstract}
	
	\begin{IEEEkeywords}
		multiobjective optimization, quality indicator, performance assessment, exact method, evolutionary algorithms
	\end{IEEEkeywords}
	
	\section{Introduction}

  	Many real-world optimization problems are composed of multiple and conflicting objectives. Although traditional approaches can be used to combine the objectives into a single one and solve the resulting problem, several multi and many-objective optimization techniques have proven to be efficient techniques dealing with the true multiobjective nature of such problems. \cite {articleCSWM}. 
	
	The solution sets are formed in such a way that each solution represents a trade-off among objectives. If a comparison of different solution sets is needed, many performance measures can be applied \cite{Li:2019:QES:3320149.3300148}. Graphical techniques represent an alternative way to help examine the solution sets visually. Those techniques are quite useful when the problems have two or three objectives only. However, when the number of objectives is higher than three, this task is challenging (if not impractical), needing proper visualization techniques that can exhibit solution set features like location, shape, and distribution \cite{DBLP:journals/swevo/IbrahimRMD18}.
		
    When it is necessary to summarize the solution sets, taking into account their characteristics and features, quality indicators are widely applied \cite{1197687}. These indicators have been used to compare the outcomes of multiobjective optimizers quantitatively. Ideally, a quality indicator should be able not only to state whether an outcome is better than others but also in what aspects. In a recent paper \cite{Li:2019:QES:3320149.3300148}, 100 quality indicators were discussed including some that are considered state-of-the-art. The quality aspects of these indicators, as well as their strengths and weaknesses, are examined and compared.
	
    A unary quality indicator is a mapping that assigns an approximation set to a real number \cite{Emmerich2018}, and it is used to compare approximation sets generated by an optimizer. Inverted generational distance (IGD) \cite{10.1007/978-3-319-15892-1_8}, hypervolume (HV) \cite{8625504}, \cite{Yang2019},\cite{Bradford_2018}, and  $\epsilon$-additive/multiplicative indicator \cite{1197687} are some examples, to name a few. Despite their applicability, some indicators require a pre-defined reference point or the knowledge of the true Pareto front.
	
   While unary indicators are able to summarize only one solution set, binary indicators take into account two approximation sets and return a real value, which can be used to say whether an approximation is better than others. For two sets $P$ and $Q$, for example, if $P$ weakly dominates $Q$, then $I(P, Q) = 0$. If P dominates some points of Q, and Q does not dominate any point of P, it is fair to expect that the indicator supports $P$ to $Q$.
	
	In \cite{DBLP:journals/corr/LiY17a}, a new quality measure, called dominance move (DoM) is proposed.  DoM measures the minimum `effort' that one solution set has to make in trying to dominate another set, more specifically the sum of the movement needed to make a set dominant. It has the same $\epsilon$-indicators' interpretation and it can capture all quality aspects of solution sets, such as Pareto convergence and spread. The authors propose an exact algorithm to calculate DoM for the bi-objective case that can be computed in low computational cost. However, as stated in \cite{DBLP:journals/corr/LiY17a}, it can not be used or extended to three or more dimensions due to the indicator combinatorial nature.  
	
	In an attempt to overcome this difficulty, this work focuses on a DoM formulation as an assignment problem, and its solution using a mixed-integer programming method \cite{DBLP:books/daglib/0023873} is proposed.  Assignment problems have some variants; however, it is common for the problems to involve a form to optimally match the elements of two or more sets, in which the assignment's complexity refers to the number of items to be matched \cite{Assign_2007}.
	
    The paper is organized as follows. In Section \ref{relatedwork}, some definitions and related work are presented. Section \ref{assignment} introduces the DoM indicator and our formulation to treat it as an assignment problem. Next, a mixed-integer programming formulation is also presented as an assignment implementation. Its strengths and weaknesses to solve the assignment problem are also discussed. Some experiments are presented in Section \ref{experiments}, firstly in the bi-objective case, showing that DoM overcomes some $\epsilon$-indicators weaknesses, and it is still in agreement with the DoM calculation algorithm developed in \cite{Li:2019:QES:3320149.3300148}. Moreover, in the experiments section, some solution sets were generated by IBEA, NSGAII, and SPEA2 and used to assess and compare the DoM indicator with other common indicators (HV and IGD). In Section \ref{conclusion}, some observations and future research considerations are finally described.
	
	\section{Definitions and related work}\label{relatedwork}
	In general, a multiobjective optimization problem (MOP) includes \textbf{\textit{x}} decision variable vector from a decision space $\Omega \subseteq R^{N}$, and a set of \textit{M} objective functions. Without loss of generality, a minimization MOP can be simply defined as \cite{Yuan2018}:
	\begin{align}\label{MOP_definition}
	Min \quad  F (\textbf{\textit{x}}) = {[f_{1}(\textbf{\textit{x}}), ... ,f_{|M|}(\textbf{\textit{x}})]}^{T},  &&  \textbf{\textit{x}} \in \Omega 
	\end{align}

	The $F: \Omega \rightarrow \Theta \subseteq R^{M}$ is formed by a set  of \textit{M} objective functions, which is a mapping from decision space $\Omega$ to vectors in \textit{M}-dimensional objective space $\Theta$. We are interested in the evaluation of these objective vector (solution) sets, and the comparison relation among these vectors.
	
	Considering two solutions \textit{\textbf{p}}, \textit{\textbf{q}}  $\in \Theta$, it is possible to establish a relation in which \textit{\textbf{p}} is said to \textit{weakly dominate} \textit{\textbf{q}} if \textbf{$p_i$} $\leq$ \textbf{$q_i$} for 1 $\leq$ i $\leq$ M, and is denoted as  \textit{\textbf{p}} $\preceq$ \textit{\textbf{q}}. In addition, if there is at least one objective $i$ in which \textbf{$p_i$} $<$ \textbf{$q_i$} then it is said that \textit{\textbf{p}} \textit{dominates} \textit{\textbf{q}}, and is denoted as \textit{\textbf{p}} $\prec$ \textit{\textbf{q}}.
	A solution \textbf{p} $\in \Theta$  is called Pareto optimal  if there
	is no \textbf{q}  $\in  \Theta$ that dominates \textbf{p}. The set of all Pareto optimal solutions of an MOP is called Pareto optimal frontier. In the same way, the weak dominance  relation can be defined to solution sets:
	
	{\textit{Weak Dominance}:} 
	The set $P$ weakly dominates $Q$, denoted as $P$ $\preceq$ $Q$, if every solution \textbf{\textit{q}} $\in $ $Q$ is weakly dominated by at least one solution \textbf{\textit{p}} $\in$ $P$ .
	
	The goal of a multiobjective algorithm is to generate approximation sets representing the Pareto front of a MOP. In the last years, the growth of multiobjective algorithms lead to a key issue: the evaluation and comparison of approximation sets generated by these algorithms. To assess the quality of sets in MOP, one must take into account several aspects, such as convergence to the true Pareto front, spread of the solution, etc.  Quality indicators represent a way to quantitatively evaluate the approximation sets generated by different algorithms. Ideally, a quality indicator should not only be able to say which algorithm is better than the other but also to identify in what aspects. 
	The following definition formalizes a quality indicator \cite{1197687}:
	
	{\textit{Quality indicator}:} 
	An \textit{k}-ary quality indicator \textit{I} is a function \textit{I}:${\Theta}^k \rightarrow {\rm I\!R}$ , which assigns each vector of \textit{k} solutions sets {($P_1, P_2, ..., P_k$)} a real value I{($P_1, P_2, ..., P_k$)}.
	
	Quality indicators can be unary, binary, or \textit{k}-ary, defining a real value to one solution set, two solution sets, or \textit{k} solution sets, respectively. For a comprehensive review of quality indicators, in \cite{Li:2019:QES:3320149.3300148}, some indicators are defined and discussed using their quality facets as: \textit{convergence}, \textit{spread}, \textit{uniformity}, and \textit{cardinality}. Issues such strengths, weaknesses, and evaluation are also analyzed. 
	
	Many indicators have been used in multiple situations in the literature \cite{Li:2019:QES:3320149.3300148}. Hypervolume (HV), used in \cite{Yang2019},\cite{Bradford_2018}, inverted generational distance (IGD) used in \cite{10.1007/978-3-319-15892-1_8}, and $\epsilon$-indicator are some examples. DoM and these indicators  will be used in this paper and are defined below:
	\begin{itemize}{}{}
		\item \textbf{Hypervolume (HV)}:  Let $r^{'}= (r_1, ..., r_m)$ be reference points in the objective space that is dominated by all approximation sets. Let $P$ be one approximation set. The HV value of $P$ with regard to $r^{'}$ represents the volume of the region which is dominated by $P$ and dominates $r^{'}$. Generally, the computational cost is 
		exponential regarding  to the number of objectives.
		
		\item \textbf{Inverted generational distance (IGD)}:  Let $R^{*}= ({r^{*}}_1, ..., {r^{*}}_m)$ be a reference set of uniformly distributed points on the Pareto front. Considering $P$ as an approximation set to the Pareto front, the inverted generational distance between $R^{*}$ and $P$ is defined as:
		\begin{equation}\label{IGD}
		IGD(R^{*}, P) = \frac{ \sum\limits_{r \in R^{*}} {d(r,P)}}{|{R^{*}}|}
		\end{equation}
		
		${d(r,P)}$ is the minimum Euclidean distance from point $r$ to approximation set $P$.  The IGD metric is able to measure both diversity and convergence of $P$ if $|{R^{*}}|$ is large enough \cite{cheng2018benchmark}. The computational cost is $O( |M|\times |R^{*}| \times |P|)$, where $|M|$ is the number of objectives.
		
		\item \textbf{$\epsilon$-additive/multiplicative indicator}: it is a extension to the evaluation of approximation schemes in operational research and theory \cite{1197687}. For two solution sets $P$ and $Q$, the additive $\epsilon$-indicator, $I_{\epsilon}(P, Q)$ is the minimum value that can be added to each solution in Q, such that they become weakly dominated by at least one solution in P. Formally, the additive $\epsilon$-indicator is calculated as:
		\begin{equation}\label{epslonindicator}
		I_{\epsilon^{+}}(P, Q) =  \max\limits_{q \in Q} \min\limits_{p \in P} \max\limits_{i \in \{1..  |M|\}} {p^{i} - q^{i}}
		\end{equation}
		in which $p^{i}$ denotes the objective value of solution $p$ in the \textit{ith} objective, and $|M|$ is the number of objectives. For the multiplicative $\epsilon$-indicator, the ${p^{i} - q^{i}}$ is replaced by $\frac{p^{i}}{q^{i}}$. 
		A value of $I_{\epsilon^{+}}(P, Q)$ $\leq$ 0 or $I_{\epsilon^{\times}}(P, Q)$ $\leq$ 1 implies that $P$ weakly dominates $Q$. The computational cost is $O(|M| \times |P| \times |Q|)$.
		
		\item \textbf{Dominance move (DoM)}: it is a measure for comparing two sets of multidimensional points being classified as a binary indicator. It considers the movement of points in one set to make this set weakly dominated by the other set. DoM can be defined as follows,\cite{DBLP:journals/corr/LiY17a}:
		Let $P$ be a set of points in ${ \{p_1,p_2,..,p_{NP} \}}$ and $Q$ be a set of points in ${ \{q_1,q_2,..,q_{NQ} \}}$. The dominance move of $P$ to $Q$, $D(P,Q)$, is the minimum total distance of moving points of P, such that any point in $Q$ is weakly dominated by at least one point in $P$. In fact, the problem is to find ${ \{{p^{'}}_1,{p^{'}}_2,..,{p^{'}}_{NP} \}}$ from  ${ \{p_1,p_2,..,p_{NP} \}}$ positions, such that $P^{'}$ weakly dominates $Q$, and the total move from ${ \{p_1,p_2,..,p_{NP} \}}$ to ${ \{{p^{'}}_1,{p^{'}}_2,..,{p^{'}}_{NP} \}}$, denoted as $d(p_i,p^{'}_i)$,  must be minimum. The formal definition of DoM can be expressed as: 
		\begin{equation}\label{DPQ}
		D(P,Q) = \underset{P^{'} \preceq  Q}{minimize} \sum\limits_{i=1}^{NP} d(p_i,p^{'}_i) 
		\end{equation}
	\end{itemize}
	
	The number of possibilities to find $P^{'}$ is numerous. Any combination of some $P^{'}$ can dominate $Q$, considering (\ref{DPQ}). %Consider $P^{'}_{(1..s)}$ from $P$ with some  movement updates in the objectives in conjunction with other $P^{'}_{(s+1..np)}$ from original $P$, for example. 
	The  authors of \cite{DBLP:journals/corr/LiY17a}, proposed an exact solution for calculating DoM in a bi-objective case \cite{Li:2019:QES:3320149.3300148}. The algorithm can be outlined as:
	\begin{list}{}{}
		\item \textit{Step 1:} Remove the dominated points in both $P$ and $Q$, separately. Remove the points of $Q$ that are dominated by at least one point in $P$.
		\item \textit{Step 2:} Denote  $R$ $=$ $P \cup Q$ and start the process. Each point of $Q$ in $R$ is considered as a group. For each group of $Q$, find its inward neighbor $\textit{\textbf{r}} = n_{R}(\textit{\textbf{q}})$ in $R$. If the point $\textbf{r} \in P$, then merge $\textit{\textbf{r}}$ into the group of $\textit{\textbf{q}}$, otherwise $\textbf{\textit{r}} \in Q$. If $\textbf{\textit{r}}$ is not assigned to one group, merge the two groups of $\textbf{\textit{q}}$ and $\textbf{\textit{r}}$ into one group.
		\item \textit{Step 3:} If there exists no point $\textbf{\textit{q}} \in Q$ such that $\textbf{\textit{q}} = n_{R}(n_{R}(\textbf{\textit{q}}))$ (i.e., there is a loop between the points) in any group, then the procedure ends and there is an optimal solution to the case.
		\item \textit{Step 4:} There is a loop in one or some groups. The procedure replaces these loops with the ideal point. The ideal point is formed by the best of each objective in each point inside the loop or group. Return to step 3 until convergence.
	\end{list}
	The definitions, theorems, and corollaries to prove that this algorithm is correct in the bi-objective case are presented in \cite{Li:2019:QES:3320149.3300148}. Furthermore, DoM is Pareto dominant compliant and any prior problem knowledge and pre-defined parameter are not necessary. However, due to the combinatorial nature of the problem, the authors stated that there is no solution for three or more objectives.

	\section{The dominance move calculation as an assignment problem}\label{assignment}
	
	Our proposal concept of DoM  calculation is based on the observation that the problem is, in fact, a particular case of an assignment problem with two levels and some constraints. To deal with the problem, we have to find an assignment of $P$ to $Q$ with the restrictions that each \textbf{\textit{q}} must be assigned to one \textbf{\textit{p}} with the minimum distance. Nevertheless, in classic assignment problems,  $P$ does not change its features, and this aspect must be considered for the DoM calculation. 
	
	A simple and hypothetical example to clarify the situation can be given as follows: consider $P$ as \{(1.5, 1.3, 1.1), (1.4, 2.1, 1.8)\} and Q as \{(1.4, 1.2, 1.0), (1.3, 2.0, 2.0)\}. The possible inward neighbor $\textit{\textbf{r}} = n_{R}(\textbf{\textit{q}})$ of points ${q_{1}}$,  and $q_{2}$ can be, respectively, $p_{1}$ and $p_{2}$. This creates an assignment of $P$ to $Q$ with the minimum $D(P,Q)$, considering that $P$ is fixed: $D(P,Q) = d(p_{1},q_{1}) + d(p_{2},q_{2})) = 0.5$. However, if we considered a movement from $P$ to $P^{'}$, then $p_1$ would be transformed into $p^{'}_1$ = \{(1.4, 1.3, 1.1)\}. In this sense, we can find a better assignment and lower value of $D(P,Q) =d(p_{1},p^{'}_{1}) + d(p^{'}_{1},q_{1}) + d(p^{'}_{1},q_{2}) = 0.4$. Clearly, other assignments from $P$ to $P^{'}$ and  to ${Q}$ are capable to generate the same value.
	
	\definecolor{myblue}{RGB}{10,70,160}
	\definecolor{myblue2}{RGB}{100,100,120}
	\definecolor{mygreen}{RGB}{80,160,80}
	\begin{figure}[!htb]
		\begin{tikzpicture}[thick,
		every node/.style={draw,circle},
		psnode/.style={fill=myblue},
		pslnode/.style={fill=myblue2},
		qsnode/.style={fill=mygreen},
		every fit/.style={ellipse,draw,inner sep=-1pt,text width=2cm},
		->,shorten >= 3pt,shorten <= 3pt
		]
		
		% the vertices of P
		\begin{scope}[start chain=going below,node distance=7mm]
		
		\foreach \i in {1,2,...,3}
		\node[psnode,on chain] (p\i) [label=left: ${p}_\i$] {};
		\end{scope}
		
		% the vertices of P_'
		\begin{scope}[xshift=3cm,start chain=going below,node distance=7mm]
		\foreach \i in {1,2,...,9}
		\node[pslnode,on chain] (pl\i) [label=left: $p^{'}_{\i}$] {};
		\end{scope}
		
		% the vertices of Q
		\begin{scope}[xshift=6cm,start chain=going below,node distance=7mm]
		\foreach \i in {1,2,...,3}
		\node[qsnode,on chain] (q\i) [label=right: $q_{\i}$] {};
		\end{scope}

		% the edges P to P'
		\foreach \j in {1,2,...,3}
		{
			\draw (p1) -- (pl\j) ;
		}
		\foreach \j in {4,5,...,6}
		{
			\draw (p2) -- (pl\j) ;
		}
		\foreach \j in {7,8,...,9}
		{
			\draw (p3) -- (pl\j) ;
		}
		
		% the edges P' to 1
		\foreach \j in {1,2,...,9}
		\foreach \k in {1,2,...,3}
		{
			\draw (pl\j) -- (q\k) ;
		}
		
		\end{tikzpicture}
		\centering
		\caption[multilevel assignment]{One possible example of assignment between $P$, with \textit{NP} = 3,  and $Q$ with \textit{NQ} = 3. Considering improvements in $P$, $P^{'}$ is generated, and in this example, $NP^{'}$ = 9. The distances between  $P$ to $P^{'}$ and $P^{'}$ to $Q$  are, respectively, $d(p_i,p^{'}_k)$  and $d(p^{'}_k,q_j)$ corresponding to edges.}
		\label{fig:sampleassignment}
	\end{figure}
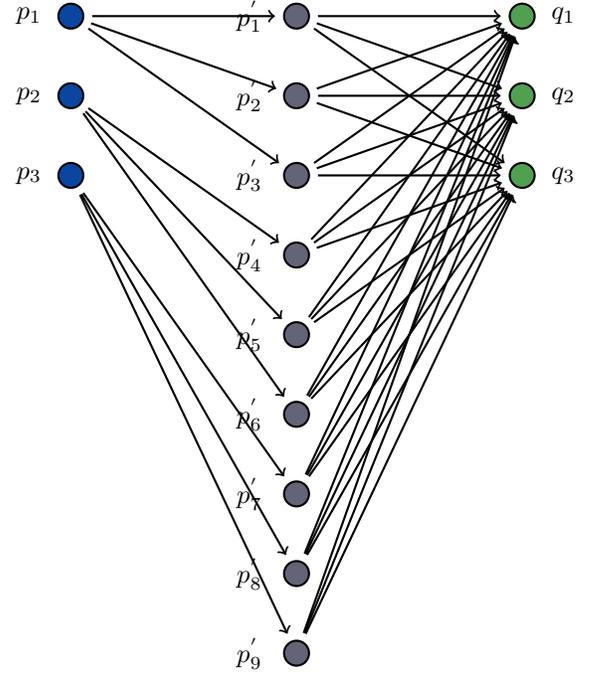
	
	Inspired by an assignment problem approach, the proposed DoM calculation is explained in detail. Let $P$ and $Q$ be two sets of points, with  $ p_i$ points $ i \in \{{1, .., NP}\}$ and $q_j$ points $j \in \{{1, .., NQ}\} $. $P$ and $Q$ are given in the problem. 
	$P^{'}$ is a set of points containing the candidates to dominate $Q$ with some update in one or many objectives. This update generates a better distance such as expressed in \eqref{DPQ}. It should be noted that $ p^{'}_k$ with $k \in \{{1, .., NP^{'}}\}$ can be generated from $p$.
	
	To better understand our approach, this concept is illustrated in Figure(\ref{fig:sampleassignment}).  Suppose the solution sets \textit{P} and \textit{Q} have both $ NP = NQ = 3$. $P^{'}$ is generated with 3 possibilities for each $p_i$. In total, there are 9 $p^{'}_k$ generated.  The first assortment of edges from  $p_i$ to $p^{'}_k$ represents the distance $d(p_i,p^{'}_k)$ as a way to improve $p_i$ generating $p^{'}_k$ candidate. The second assortment is from $p^{'}_k$ to $q_j$, and represents the distance $d(p^{'}_k,q_j)$, which can be seen as the distance from some $p^{'}_k$ to weakly dominate some $q_j$ or a $g$ group formed by more than one $q_j$.
	
	In a typical assignment problem, the goal is to find a one-to-one match between \textit{n} tasks and \textit{o} agents, for example. The objective function minimizes the total cost of the assignments as $c_{(i,j)}$, from task $j$ to agent $i$. At most, one agent must do a task, and every task must be done, as proposed in \cite{Assign_2007}. The mathematical model for the classic assignment problem is given as in \eqref{mip_classic}:
	\begin{align}\label{mip_classic}
	& \underset{}{\text{minimize}}
	\sum\limits_{i=1}^{N} \sum\limits_{j=1}^{O} c_{(i,j)}  x_{(i,j)}  \nonumber\\
	& \text{subject to} \nonumber \\
	& \sum\limits_{i=1}^{N}x_{(i,j)} = 1 ,  \quad  \forall \mathit{j} \in {O} \nonumber &\\
	& \sum\limits_{j=1}^{O}x_{(i,j)} = 1 ,  \quad  \forall \mathit{i} \in {N} \nonumber &\\
	& x_{(i,j)} \in \{0,1\}  \quad \forall \mathit{i} \in {N},  \forall \mathit{j} \in {O}
	\end{align}
	
	The DoM assignment model is detailed in \eqref{mip_dom_model_comb}. The objective function searches for a valid path from $p_{i}$ to $q_{j}$ through $p^{'}_k$ with the minimum distance between pairs. The first set of constraints guarantee that, for each $q_{j}$, there is a valid path. The next constraints involving $xc_{(k,j)}$, a  binary variable, guarantee that the path from $p_{i}$ to $q_{j}$ is valid. When $xc_{(k,j)}$ is 1, then there is a valid path in the assignment graph. On the other hand, if $xc_{(k,j)}$ is 0, the path is infeasible. Essentially, it is necessary that $p^{'}_k$, generated from $p_{i}$, must be shared between $p_{i}$ and $q_{j}$. Again, the problem can be viewed as a bipartite graph with two layers, such as the example in Figure \ref{fig:sampleassignment}.
	\begin{align}\label{mip_dom_model_comb}
	& \underset{}{\text{minimize}}
	\sum\limits_{i=1}^{NP} \sum\limits_{k=1}^{NP^{'}} d(p_i,p^{'}_k)  x_{(i,k)} +  \sum\limits_{k=1}^{NP^{'}} \sum\limits_{j=1}^{NQ} d(p^{'}_k,q_{j})  x_{(k,j)}  & \nonumber\\
	& \text{subject to} \nonumber \\
	& \sum\limits_{k=1}^{NP^{'}}xc_{(k,j)} = 1 ,  \quad  \mathit{j}= (1,..,{NQ}) \nonumber &\\
	& xc_{(k,j)}\leq  x_{(i,k)} , \quad  \mathit{i}= (1,..,{NP}), \mathit{k} =(1,..,{NP^{'}}), \nonumber \\ 
	&\quad\quad  \mathit{j}= (1,..,{NQ}) \nonumber  \\
	& xc_{(k,j)}\leq  x_{(k,j)} ,\quad \mathit{k} =(1,..,{NP^{'}}), \mathit{j}= (1,..,{NQ}) \nonumber  \\
	& xc_{(k,j)}\geq  x_{(i,k)}  + x_{(k,j)} - 1,\quad \mathit{i}= (1,..,{NP}), & \nonumber \\  
	& \quad\quad  \mathit{k} =(1,..,{NP^{'}}), \mathit{j}= (1,..,{NQ}) \nonumber  \\
	& x_{(i,k)} \in \{0,1\}, \quad  \mathit{i}= (1,..,{NP}), \mathit{k} =(1,..,{NP^{'}}) \nonumber\\
	& x_{(k,j)} \in \{0,1\}, \quad \mathit{k} =(1,..,{NP^{'}}) , \mathit{j}= (1,..,{NQ})   \nonumber \\
	& xc_{(k,j)} \in \{0,1\}  \quad \mathit{k} =(1,..,{NP^{'}}), \mathit{j}= (1,..,{NQ}) 
	\end{align}
	
	In (\ref{mip_dom_model_comb}), the $d(p_i,p^{'}_k)$ and $d(p^{'}_k,q_{j})$ must be computed beforehand.  Two distance matrices can represent these two parameters. Consider that $|L|$ is the number of solutions in an arbitrary set, for example, and \textit{|M|} the number of objectives. The total number of pairwise comparisons to calculate the distance matrix is $\lceil |L|(|L| -1)/2  \rceil $. Each comparison can be a vector operation with \textit{|M|} summations to obtain a pairwise distance element. Some works deal with the task of how to calculate the distance matrices efficiently such as in \cite{Distance_Matrix_2014}. However, these computations can become prohibitive when either $|L|$ or $|M|$ are large (thousands of magnitude). For the solution sets context, the number of elements represents a critical value to be chosen.    
	
	It is important to note that model \eqref{mip_dom_model_comb} does not deal with the problem of finding the  $p^{'}_k$ candidates. Proposing the $p^{'}_k$ candidates is a hard task given  its combinatorial  nature. Still, it is possible to use the full combinatorial approach, which represents all the combinations selecting one $p_{i}$ and all the possible $g$ groups in $Q$. Another exploratory possibility could try to learn `good' candidate features. A machine learning approach could use a limited number of generated candidates and find such characteristics in candidates using a loss function as in \eqref{mip_dom_model_comb}. Considering $g$ as a group with one or many $q_{j}$, and assuming that $p_{i}$ will be used as a base to be updated, one could generate  $p^{'}_k$ candidates which weakly dominate all $g$ group while minimizing \eqref{mip_dom_model_comb}.

	\section{Experiments}\label{experiments}
	
	\subsection{Bi-objective experiments}
	
\begin{figure}[htbp]
		\centerline{\includegraphics[width=\linewidth,height=6cm]{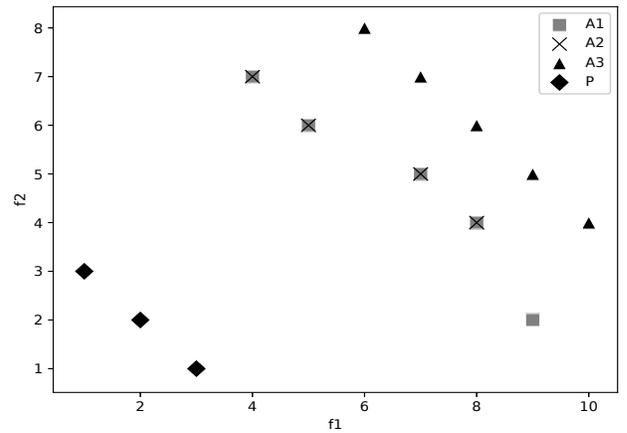}}
		\caption{Experiment  using four solution sets proposed in \cite{1197687} to show how $\epsilon$-indicators assess the sets characteristics. There are four solutions sets. P is the  Pareto front and there are explicit relations among the other solutions sets: A1 $\succeq$ A2,A1 $\succeq$ A3,A2 $\succeq$ A3}
		\label{fig:test_set_ziztler}
	\end{figure}

	The first experiment was done to show that DoM and $\epsilon$-indicators have a  similar interpretation. We used the same simple bi-objective problem proposed in \cite{1197687}. There are four solution sets as can be viewed in Figure \ref{fig:test_set_ziztler}. P is the Pareto front and there is a dominance relation among $A_{1}$, $A_{2}$ and $A_{3}$: $A_{1}$ $\succeq$ $A_{2}$,$A_{1}$ $\succeq$ $A_{3}$, $A_{2}$ $\succeq$ $A_{3}$.
	
	 It is expected that an indicator should reflect all the solution set features. In this sense, Table \ref{tab:epslon_DoM} presents the values for all combinations among  $A_{1}$, $A_{2}$, $A_{3}$, and $P$. It can be observed that DoM and $\epsilon$-indicators have the same interpretation, and the comparisons lead to the same conclusions among the solution sets. Nonetheless, it is relevant to observe some differences: 
	 \begin{itemize}
	 	\item $\epsilon$-indicators are only related to one particular solution and only one objective in whole solution set. There is  an information loss, because the indicator ignores the difference in other objectives. It can be viewed, for example, in comparison with $\epsilon$-indicator($A_{1}$,P) and DoM($A_{1}$,P). Considering $\epsilon$-indicator($A_{1}$,P), it was obtained using the first solution from \textit{A1} and \textit{P} on \textit{f2} objective; otherwise, the DoM($A_{1}$,P) has explored \textit{f1} and \textit{f2} objectives, the distance value was generated using the optimal problem resolution that was obtained from the first \textit{A1} solution $a_{1}=(4,7)$, generate a surrogate point $a'_{1}=(2,2)$ that dominate $p_{2}$ and have a  dominance move distance of one for each  $p_{1}$ and $p_{3}$, summing the whole dominance move equals to nine;
	 	\item $\epsilon$-additive is not able to capture differences concerning cardinality of solution sets (observe $\epsilon$-additive($A_{3}$,$A_{1}$) and $\epsilon$-additive($A_{3}$,$A_{2}$)). At the same time, $\epsilon$-multiplicative presents the same proportion related as DoM;
	 	\item DoM presents greater values than $\epsilon$-indicators (observe  DoM($A_{1}$,P) versus $\epsilon$-indicator($A_{1}$,P), or DoM($A_{3}$,P) versus $\epsilon$-indicator($A_{3}$,P)). This fact can be explained since DoM takes into account information from all objectives.
	 \end{itemize}
	
	The $\epsilon$-indicators also measure the minimum value added to one solution set to make it be weakly dominated by another set. However, as it can be observed in Table \ref{tab:epslon_DoM}, there is an information loss. This information loss is critical, considering many objectives scenarios. One simple example, proposed in \cite{DBLP:journals/corr/LiY17a}, can be easily observed: consider two 10-objective solutions, such as $p_{1}=\{0,0,0,..,1\}$ and $q_{1}=\{1,1,1,..,0\}$. In this case,  $\epsilon$-additive($p_{1}$, $q_{1}$) =  $\epsilon$-additive($q_{1}$, $p_{1}$) = 1.
	
	The second experiment was done to show the correctness of DoM assignment calculation, and how it addresses the quality indicator facets: convergence, spread, uniformity, and cardinality  \cite{Li:2019:QES:3320149.3300148}. The same guidelines proposed in \cite{DBLP:journals/corr/LiY17a} to solve DoM in the bi-objective case were applied. The data was provided by Dr Miqing Li. Our method presented the same results, which were found in the original work. This concordance showed that the proposed DoM assignment model was not only correct, but in agreement with the DoM concept and with the exact algorithm for the bi-objective case presented in \cite{DBLP:journals/corr/LiY17a}.

	\begin{table}[htbp]
		\caption{Comparisons amongst $\epsilon$-additive, $\epsilon$-multiplicative, and DoM  indicators.	A value of $\epsilon$-additive $\leq$ 0, $\epsilon$-multiplicative $\leq$ 1 or DoM $\leq$ 0  implies that P weakly dominates Q. The solutions sets are presented graphically in Figure \ref{fig:test_set_ziztler}}
		\begin{center}
			\begin{tabular}{|c|c|c|c|c|c|}
				\hline
				\textbf{Quality}&\textbf{P solution}&\multicolumn{4}{|c|}{\textbf{Q solution sets}} \\
				\cline{3-6} 
				\textbf{indicator}&\textbf{sets}  & $A_{1}$& $A_{2}$&$A_{3}$&$P$\\
				\cline{1-6} 
				 &$A_{1}$&0.000&0.000&-1.000&4.000\\
				\cline{2-6} 
			$\epsilon$-additive	&$A_{2}$&2.000&0.000&0.000&4.000\\
				\cline{2-6} 
				&$A_{3}$&2.000&2.000&0.000&5.000\\
				\cline{2-6} 	
				&$P$&-1.000&-3.000&-3.000&0.000\\
				
				\cline{1-6} 
				 &$A_{1}$&1.000&1.000&0.900&4.000\\
				\cline{2-6} 
			$\epsilon$-multiplicative	&$A_{2}$&2.000&1.000&1.000&4.000\\
				\cline{2-6} 
				&$A_{3}$&2.000&1.500&1.000&6.000\\
				\cline{2-6} 	
				&$P$&0.500&0.428&0.333&1.000\\
				
				\cline{1-6} 
				 &$A_{1}$&0.000&0.000&0.000&9.000\\
				\cline{2-6} 
			DoM	&$A_{2}$&2.000&0.000&0.000&9.000\\
				\cline{2-6} 
				&$A_{3}$&8.000&6.000&0.000&12.000\\
				\cline{2-6} 	
				&$P$&0.000&0.000&0.000&0.000\\
				\hline			
				\multicolumn{5}{l}{}
			\end{tabular}
			\label{tab:epslon_DoM}
		\end{center}
	\end{table}

	\subsection{Multiobjective experiments}
	
	After using some artificial test sets, the next experiment aimed to (i) validate the DoM assignment model using problems with three objectives and (ii) assess the comparison results with other state-of-the-art quality indicators, such as HV and IGD.  Visualization of approximation sets was also applied to provide an important insight into the properties of the approximation sets while validating the conclusions. 
	
	In all tests, algorithms such as IBEA, NSGAII, and SPEA2 were used to generate the solution sets. It is important to note  that any other algorithm could have been applied to generate the solution sets. Our main goal  was to validate the effectiveness of the proposed DoM assignment formulation and not perform an algorithm ranking. 
	
    In each experiment, and for our purpose, an important parameter had to be chosen beforehand: the definition of the population size (i.e., others parameters were kept default in each software used). The question is closely related to the $p^{'}_k$ candidates and the solution set cardinality (one of the quality indicator facet). Generally speaking, in order to have a good approximation set of the Pareto front, in terms of convergence, spread, and uniformity, the number of non dominated solutions grows exponentially concerning the problem dimension. 
	
	Using the model in \eqref{mip_dom_model_comb}, the selection of the $p^{'}_k$ candidates was done using the full combinatorial approach: all possible combinations selecting one $p_{i}$ and all the possible $g$ groups in $Q$. The number of such candidates is detailed in (\ref{comb_plinha}), in which $g$ is a group with one or many $q_{j}$, and assuming that $p_{i}$ will be used as a base to be updated, generating $p^{'}_k$, which can weakly dominate all $g$ group. 
	\begin{equation}\label{comb_plinha}
	\begin{aligned}
	& NP \sum\limits_{g=1}^{NQ}{NQ\choose g} = \\
	& NP \left({NQ\choose 1} + {NQ\choose 2} + ... + {NQ\choose NQ}\right) =  \\
	& NP( 2^{NQ} - 1)
	\end{aligned}
	\end{equation}
	
	Based on \eqref{mip_dom_model_comb} and \eqref{comb_plinha}, the population size for the algorithms was set to 20. It is relevant to note that the number of objectives does not change the model parameters, since the matrices with $d(p_i,p^{'}_k)$ and $d(p^{'}_k,q_{j})$ do not suffer structural impact (i.e., the number of row and columns remains the same, regardless of the number of objectives). 
	
	\begin{figure}[htbp]
		\centerline{\includegraphics[width=\linewidth,height=11.42cm]{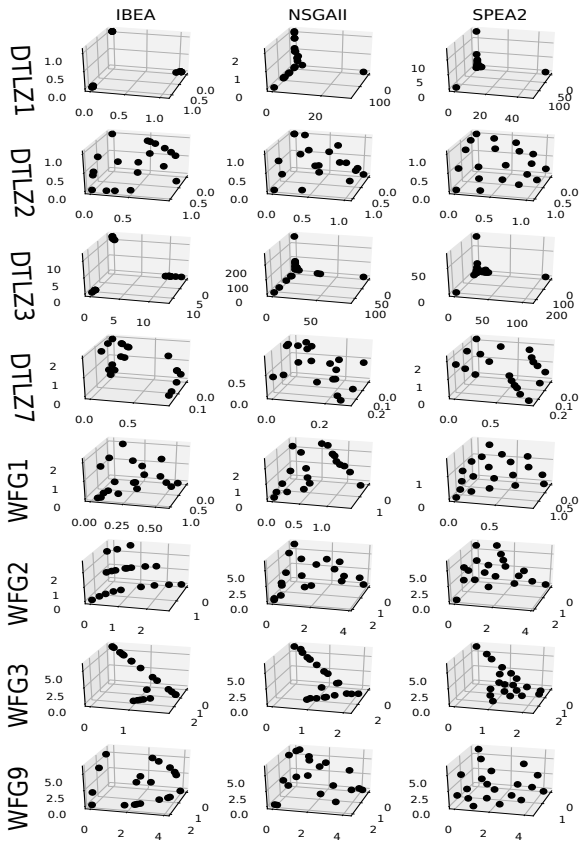}}
		\caption{Solution sets with \textit{NP} = \textit{NQ} = 20 solutions and three objectives, \textit{M} = 3, generated by IBEA, NSGAII, and SPEA2 algorithms applied to DTLZ1, DTLZ2, DTLZ3, WFG1, WFG2, WFG3 and WFG9 problem sets.}
		\label{fig:problem_set_20_3}
	\end{figure}
	
	Figure \ref{fig:problem_set_20_3} presents all the solution sets. The \textit{DTLZ} and \textit{WFG} families were chosen. These problems present many different characteristics as convexity/concavity, disconnection, multimodality, and degeneracy. In this way, DTLZ1, DTLZ2, DTLZ3, and DTLZ7, WFG1, WFG2, WFG3, and WFG9 were selected. Some algorithms were also used to test and compare the DoM with other indicators: IBEA \cite{Zitzler04indicator-basedselection}, SPEA2 \cite{Zitzler01spea2:improving}, and NSGAII \cite{Bradford_2018}. The algorithms were run 21 times using each problem set, and the final combined Pareto front is presented in \ref{fig:problem_set_20_3}. The number of function evaluations to each algorithm run was set to 10000 in each experiment.

	Tables \ref{tab:IGD} and \ref{tab:HV} show two unary quality indicators: the inverted generational distance (IGD) and hypervolume (HV) for the problem sets depicted in \ref{fig:problem_set_20_3}. It is mandatory to have a reference set and point, respectively, to calculate these indicators, and this task is not only a challenging one \cite{Li:2019:QES:3320149.3300148}, but also sometimes provided by the user \cite{Yang2019}. 
	For HV, we decided to use a dominated point chosen amongst all algorithm solutions and objectives. Concerning IGD, we built a reference set using the following strategy: we took all the solutions from all algorithms, and extracted the non dominated points, thus creating the reference set.
	
	The DoM quality measure is a binary indicator that takes as input, two approximation sets. In Table \ref{tab:Dom_values}, for example, it is possible to see all the comparisons among the solution sets generated by the algorithms for the \textit{DTLZ} and \textit{WFG} families. $P$ was the solution set generated by the algorithm which it was trying to dominate, and $Q$ was the solution set which was being dominated.
	
	\begin{table}[htbp]
		\caption{IGD quality indicator for problem sets: DTLZ1, DTLZ2, DTLZ3, DTLZ7, WFG1, WFG2, WFG3 and WFG9 generated using IBEA, NSGAII and SPEA.}
		\begin{center}
			\begin{tabular}{|c|c|c|c|}
				\hline
				\textbf{Problem}&\multicolumn{3}{|c|}{\textbf{IGD}} \\
				\cline{2-4} 
				\textbf{set} & \textbf{\textit{IBEA}}& \textbf{\textit{NSGAII}}& \textbf{\textit{SPEA2}} \\
				\hline
				DTLZ1&0.175&0.024&0.056\\
				\hline
				DTLZ2&0.091&0.091&0.078\\
				\hline
				DTLZ3&0.132&0.052&0.049\\
				\hline	
				DTLZ7&0.114&0.253&0.082\\
				\hline			
				WFG1&0.144&0.112&0.096\\
				\hline
				WFG2& 0.133&0.090&0.086\\
				\hline
				WFG3&0.061&0.055&0.053\\
				\hline	
				WFG9&0.099&0.178&0.103\\
				\hline			
				\multicolumn{4}{l}{}
			\end{tabular}
			\label{tab:IGD}
		\end{center}
	\end{table}
	
	For the DTLZ1 test set, detailed in Tables  \ref{tab:IGD} and \ref{tab:HV}, the algorithm which presented the best IGD was NSGAII. For the HV indicator, it was difficult to compare the algorithms due to the inflated values. There was a tie between IBEA and NSGAII; however, SPEA2 showed a better value. Using DoM approach and the comparison among algorithms, the sets generated by IBEA and SPEA2 were both indicated as the best solutions. The results are presented in Table \ref{tab:Dom_values}. It is  clear that DoM indicated IBEA as the best choice when compared with SPEA2, DoM(IBEA, SPEA2) = 0.769 against DoM(SPEA2, IBEA) = 1.085. The solution sets from other algorithms easily dominated NSGAII. Taking a closer look at the DTLZ1 problem set presented in Figure \ref{fig:problem_set_20_3}, IBEA showed the smallest scale in all graph axis. DoM is sensible to all objectives, and the other algorithms had points near the IBEA solution set. However, the `effort' to dominate the solution sets was smaller, favouring IBEA. 
		
	In the DTLZ2 case, IGD  did not indicate differences between IBEA and NSGAII, and, in the end the best solution set was generated by SPEA2 (see Table  \ref{tab:IGD}). Considering HV,  the best algorithms were SPEA2 and IBEA (observe Table  \ref{tab:HV}). Looking at Table \ref{tab:Dom_values}, the best values pointed to NSGAII and SPEA2. Observing NSGAII and SPEA2 in Figure \ref{fig:problem_set_20_3}, it is possible to note that there is a similar graph scale, but  SPEA2 and NSGAII presented a better uniformity among the points in each solution set.
	
    The results for DTLZ3 were presented in Tables  \ref{tab:IGD} and \ref{tab:HV}: for IGD, the best algorithm was SPEA2; and for HV, SPEA2 had the best value. However, it is relevant to note that these problem sets showed inflated solutions in the same way as DTLZ1. The DoM values among all algorithms (as shown in Table \ref{tab:Dom_values}) favoured IBEA and SPEA2, but in a two by two comparison, SPEA2 had a better value when compared to IBEA, DoM(SPEA2, IBEA) = 7.645. 
	
	 In the DTLZ7 problem set, the best HV values were given by IBEA and SPEA2. Considering IGD, the best one was for SPEA2. Using Table \ref{tab:Dom_values}, SPEA2 generated the best candidate solutions. Again, the values were smaller, when compared to DoM values from SPEA2 to dominate all other sets.
		
	\begin{table}[htbp]
		\caption{Hypervolume quality indicator for problem sets: DTLZ1, DTLZ2, DTLZ3, DTLZ7, WFG1, WFG2, WFG3 and WFG9 generated using IBEA, NSGAII and SPEA.}
		\begin{center}
			\begin{tabular}{|c|c|c|c|}
				\hline
				\textbf{Problem}&\multicolumn{3}{|c|}{\textbf{HV}} \\
				\cline{2-4} 
				\textbf{set} & \textbf{\textit{IBEA}}& \textbf{\textit{NSGAII}}& \textbf{\textit{SPEA2}} \\
				\hline
				DTLZ1&1.048e+05&1.048e+05&1.049e+05\\
				\hline
				DTLZ2 &0.352& 0.319&0.356\\
				\hline
				DTLZ3 &5.976e+06&5.969e+06&5.977e+06\\
				\hline	
				DTLZ7 &0.138&0.108&0.124\\
				\hline			
				WFG1&3.832&3.255&3.541\\
				\hline
				WFG2&39.543&37.168&38.626\\
				\hline
				WFG3&17.952&16.757&16.492\\
				\hline	
				WFG9& 25.798&13.992&20.052\\
				\hline			
				\multicolumn{4}{l}{}
			\end{tabular}
			\label{tab:HV}
		\end{center}
	\end{table}
	
	In the WFG family, results are presented in Tables \ref{tab:IGD} and \ref{tab:HV}. The best algorithms were SPEA2 and NSGAII for IGD; however, for HV indicator, the best one was IBEA. In Table \ref{tab:Dom_values}, the best algorithm was IBEA. Comparing IBEA to SPEA2, for example, IBEA had a lower value of DoM, DoM(IBEA, SPEA2) = 0.493, in contrast with  DoM(SPEA2, IBEA) = 0.655. Observe that the values were close to each other. 
	
	In WFG2, the best values for IGD, SPEA2 and NSGAII, were close, presenting a little difference (see Table \ref{tab:IGD}). Considering HV, the best one was the IBEA  (Table \ref{tab:HV}), but the values were once again subtle. Using DoM, detailed in Table \ref{tab:Dom_values}, there is an indication that IBEA was the best one when comparing the algorithms in a two-by-two manner. Something that should be noted is that the values were close to each other in DoM; the same phenomenon could be observed in IGD and HV, as well.  
	
    Using WFG3, for the IGD indicator, SPEA2 was the best one (subtle difference related to NSGAII), and IBEA was the best solution set considering HV.  Assessing DoM in Table \ref{tab:Dom_values}, there was an indication that IBEA also had better values.
    
    Finally, for the WFG9 problem set, Tables \ref{tab:IGD} and \ref{tab:HV} showed that for IGD,  IBEA had lower value. For HV, IBEA algorithm had a better value. Considering DoM, presented in Table \ref{tab:Dom_values}, IBEA was clearly the most competitive algorithm presenting the best values.

	\begin{table}[htbp]
		\caption{DoM values for some members of the problem set families \textit{DTLZ} and \textit{WFG}. The approximation sets were generated by IBEA, NSGAII, and SPEA2 algorithms. It must be noted that \textit{\textbf{P}} was the solution set generated by the algorithm that it was trying to dominate, and \textit{\textbf{Q} } was the solution set generated by the algorithm being dominated. Each solution set had \textit{NP} = \textit{NQ} = 20 solutions and \textit{M} = 3  objectives.}
		\begin{center}
			\begin{tabular}{|c|c|c|c|c|}
				\hline
				\textbf{}&\multicolumn{4}{|c|}{\textit{\textbf{DoM(P,Q)}}} \\
				\cline{2-5} 
				\textbf{Problem set} & & \multicolumn{3}{|c|}{\textit{\textbf{Q}}} \\
				\cline{3-5} 
				&\textbf{\textit{P}} & \textbf{\textit{IBEA}}& \textbf{\textit{NSGAII}}& \textbf{\textit{SPEA2}} \\
				\hline	
				&IBEA&0.000&0.121&0.769\\
				{DTLZ1}&NSGAII&1.535&0.000&1.535\\
				&SPEA2&1.085&0.176&0.000\\				
				\hline
				&IBEA&0.000&0.929&0.966\\
				{DTLZ2}&NSGAII&0.757&0.000&0.778\\
				&SPEA2&0.726&0.908&0.000\\				
				\hline
				&IBEA&0.000&0.044&8.137\\
				{DTLZ3}&NSGAII&16.468&0.000&16.474\\
				&SPEA2&7.645&2.121&0.000\\				
				\hline
				&IBEA&0.000&0.417&0.721\\			
				{DTLZ7}&NSGAII&1.559&0.000&1.736\\
				&SPEA2&0.667&0.420&0.000\\				
				\hline
				&IBEA&0.000&0.839&0.493\\
				{WFG1}&NSGAII&1.092&0.000&1.015\\
				&SPEA2&0.655&1.506&0.000\\				
				\hline
				&IBEA&0.000&1.393&1.401\\
				{WFG2}&NSGAII&1.514&0.000&1.608\\
				&SPEA2&1.556&1.731&0.000\\				
				\hline
				&IBEA&0.000&1.602&1.440\\
				{WFG3}&NSGAII&2.226&0.000&1.923\\
				&SPEA2&2.732&1.954&0.000\\				
				\hline
				&IBEA&0.000&1.491&1.442\\
				{WFG9}
				&NSGAII&2.643&0.000&2.372\\
				&SPEA2&3.210&2.394&0.000\\				
				\hline
				\multicolumn{4}{l}{}
			\end{tabular}
			\label{tab:Dom_values}
		\end{center}
	\end{table}

	\begin{table}[htbp]
		\caption{Experiment execution metrics: simplex iterations (from branch and bound execution), and the time spent in seconds to solve the model. Descriptive statistics are minimum, maximum, and quartiles. }
		\begin{center}
			\begin{tabular}{|c|c|c|c|c|c|}
				\hline
				\textbf{}&\multicolumn{5}{|c|}{\textbf{Descriptive Statistics}} \\
				\cline{2-6} 
				\textbf{Metric} & \textbf{\textit{Min}}& \textbf{\textit{Q1}}& \textbf{\textit{Q2}} & \textbf{\textit{Q3}}& \textbf{\textit{Max}}\\
				\hline
				simplex iterations&26212&43822&60457&75103&163869\\
				\hline
				time(seconds)&2.000  &7.475&17.595&42.255&98.550  \\
				\hline
				\multicolumn{6}{l}{}
			\end{tabular}
			\label{tab:metrics_execution}
		\end{center}
	\end{table}
	
	All the experiments were done using \textit{Platypus} \cite{Brockhoff:2019:Platypus} and \textit{PyGMO} \cite{PyGMO} to generate the problem sets and to calculate the indicators (HV and IGD). The model in  \eqref{mip_dom_model_comb}  was implemented using \textit{Python} and \textit{GUROBI} \cite{gurobi} version 9.0.0 build v9.0.0rc2 running on a Linux 64 bits operational system with 12 CPU's and 16Gb of RAM. The gap solver parameter was kept as \textit{GUROBI} default value $1e-4$. 
	
	The method proposed had two stages. The first one was to calculate matrices involving the distances $d(p_i,p^{'}_k)$ and $d(p^{'}_k,q_{j})$. The distance calculation matrices were implemented in $O(n^{2})$, and, as discussed before, there is room for improvement in this implementation. The second step was to generate and solve model \ref{mip_dom_model_comb}, which was implemented using mixed-integer programming capabilities, such as branch and bound. 
	
	Descriptive statistics from the tests are presented in Table \ref{tab:metrics_execution}. There are two metrics: simplex iterations from the branch and bound algorithm, and the time spent to solve the model. The median time spent by the model was $\sim$17 seconds, with 60457 simplex iterations. In some cases, the model was solved in two seconds; however, in the worst case, the model spent $\sim$98 seconds to solve (i.e., this case happened in the WFG9 problem set when NSGAII was trying to dominate IBEA).
	
    In this section, the goal was to verify if the DoM assignment formulation was a feasible approach for dealing with problems that have three objective functions. It is worthy to note that the maximum number of points was established to 20 (more solutions in each set increase the computational complexity and time).  Additionally, it is relevant to observe that the assignment problem formulation is not affected by the problem set dimensionality. The distance matrices, which are parameters of the model, are not altered with the problem dimensionality. Moreover, the proposed method has two stages, and just the first one, distance matrices calculation, is affected by the number of objectives/dimensions, which remains viable, at least in some hundreds of objectives/dimensions.

	\section{Conclusion}\label{conclusion}
	
    DoM is a binary indicator that considers the minimum move of one set to dominate the other set weakly. The indicator is Pareto compliant and does not demand any parameters or reference sets. Besides, it treats some weaknesses which come from the $\epsilon$-indicators but offers a similar interpretation. In this sense, it represents a  natural and intuitive relation when comparing solutions, providing a valid measure to infer Pareto dominance relations, mainly in high dimensions. The great question about DoM is its calculation concerning its computational complexity.
  
    We explored a new formulation to calculate DoM and dealt with it as an assignment problem. The idea used  $P$ and $Q$, for example,  as solution sets that have to be the solutions assigned to each other.  Comparisons with artificial bi-dimensional examples were made, detailing that DoM has the same interpretation as the $\epsilon$-indicators, and that our formulation presented the same results provided by the original DoM formulation. Additionally, some problem sets in three dimensions were also tested and showed that DoM assignment results obtained were in agreement and compliant when compared with other common indicators used in literature (IGD and HV). 
	
	DoM formulations as an assignment problem brought some particular constraints and questions, as it was discussed in model formulation.  Two calculation stages were presented: i) the matrices distance calculation, which is smoothly affected by the number of objectives (e.g., for some thousands of dimensions), and ii) the  model, as an assignment formulation, implemented using mixed-integer programming, which is affected by the number of the elements in each solution sets. 
	To the best of our knowledge, even with these limitations, an exact method to calculate DoM in three or more dimensions is not known until now.
	
	As a future research, the assignment formulation could be extended. One possible idea is to introduce the distance calculation inside the mixed-integer programming model. Possibly, it could deal with a greater number of solutions in each set. Yet, another possibility is to not use a full combinatorial approach. Otherwise, a machine learning approach could be applied to learn a function that describes features that good $p^{'}$ solutions should have to dominate some $q's$ being generated by $p$ set. 
	
	Finally, DoM is an indicator that is capable of expressing many quality indicators characteristics. An indicator with such a feature could improve not only the comparison among algorithms, but also it the  strategies used by the algorithms which are indicator based, for example.

	\bibliographystyle{IEEEtran}
	\bibliography{IEEEabrv,WCCI_DOM}

% Generated by IEEEtran.bst, version: 1.14 (2015/08/26)
\begin{thebibliography}{10}
\providecommand{\url}[1]{#1}
\csname url@samestyle\endcsname
\providecommand{\newblock}{\relax}
\providecommand{\bibinfo}[2]{#2}
\providecommand{\BIBentrySTDinterwordspacing}{\spaceskip=0pt\relax}
\providecommand{\BIBentryALTinterwordstretchfactor}{4}
\providecommand{\BIBentryALTinterwordspacing}{\spaceskip=\fontdimen2\font plus
\BIBentryALTinterwordstretchfactor\fontdimen3\font minus
  \fontdimen4\font\relax}
\providecommand{\BIBforeignlanguage}[2]{{%
\expandafter\ifx\csname l@#1\endcsname\relax
\typeout{** WARNING: IEEEtran.bst: No hyphenation pattern has been}%
\typeout{** loaded for the language `#1'. Using the pattern for}%
\typeout{** the default language instead.}%
\else
\language=\csname l@#1\endcsname
\fi
#2}}
\providecommand{\BIBdecl}{\relax}
\BIBdecl

\bibitem{articleCSWM}
S.~Chand and M.~Wagner, ``Evolutionary many-objective optimization: A
  quick-start guide,'' \emph{Surveys in Operations Research and Management
  Science}, vol.~20, pp. 35--42, 12 2015.

\bibitem{Li:2019:QES:3320149.3300148}
\BIBentryALTinterwordspacing
M.~Li and X.~Yao, ``Quality evaluation of solution sets in multiobjective
  optimisation: A survey,'' \emph{ACM Comput. Surv.}, vol.~52, no.~2, pp.
  26:1--26:38, Mar. 2019. [Online]. Available:
  \url{http://doi.acm.org/10.1145/3300148}
\BIBentrySTDinterwordspacing

\bibitem{DBLP:journals/swevo/IbrahimRMD18}
\BIBentryALTinterwordspacing
A.~Ibrahim, S.~Rahnamayan, M.~V. Martin, and K.~Deb, ``3d-radvis antenna:
  Visualization and performance measure for many-objective optimization,''
  \emph{Swarm and Evolutionary Computation}, vol.~39, pp. 157--176, 2018.
  [Online]. Available: \url{https://doi.org/10.1016/j.swevo.2017.09.011}
\BIBentrySTDinterwordspacing

\bibitem{1197687}
E.~{Zitzler}, L.~{Thiele}, M.~{Laumanns}, C.~M. {Fonseca}, and V.~G. {da
  Fonseca}, ``Performance assessment of multiobjective optimizers: an analysis
  and review,'' \emph{IEEE Transactions on Evolutionary Computation}, vol.~7,
  no.~2, pp. 117--132, April 2003.

\bibitem{Emmerich2018}
\BIBentryALTinterwordspacing
M.~T.~M. Emmerich and A.~H. Deutz, ``A tutorial on multiobjective optimization:
  fundamentals and evolutionary methods,'' \emph{Natural Computing}, vol.~17,
  no.~3, pp. 585--609, Sep 2018. [Online]. Available:
  \url{https://doi.org/10.1007/s11047-018-9685-y}
\BIBentrySTDinterwordspacing

\bibitem{10.1007/978-3-319-15892-1_8}
H.~Ishibuchi, H.~Masuda, Y.~Tanigaki, and Y.~Nojima, ``Modified distance
  calculation in generational distance and inverted generational distance,'' in
  \emph{Evolutionary Multi-Criterion Optimization}, A.~Gaspar-Cunha,
  C.~Henggeler~Antunes, and C.~C. Coello, Eds.\hskip 1em plus 0.5em minus
  0.4em\relax Cham: Springer International Publishing, 2015, pp. 110--125.

\bibitem{8625504}
J.~{Deng} and Q.~{Zhang}, ``Approximating hypervolume and hypervolume
  contributions using polar coordinate,'' \emph{IEEE Transactions on
  Evolutionary Computation}, vol.~23, no.~5, pp. 913--918, Oct 2019.

\bibitem{Yang2019}
\BIBentryALTinterwordspacing
K.~Yang, M.~Emmerich, A.~Deutz, and T.~B{\"a}ck, ``Efficient computation of
  expected hypervolume improvement using box decomposition algorithms,''
  \emph{Journal of Global Optimization}, vol.~75, no.~1, pp. 3--34, Sep 2019.
  [Online]. Available: \url{https://doi.org/10.1007/s10898-019-00798-7}
\BIBentrySTDinterwordspacing

\bibitem{Bradford_2018}
E.~Bradford, A.~Schweidtmann, and A.~Lapkin, ``Efficient multiobjective
  optimization employing gaussian processes, spectral sampling and a genetic
  algorithm,'' \emph{Journal of Global Optimization}, vol.~71, 02 2018.

\bibitem{DBLP:journals/corr/LiY17a}
\BIBentryALTinterwordspacing
M.~Li and X.~Yao, ``Dominance move: {A} measure of comparing solution sets in
  multiobjective optimization,'' \emph{CoRR}, vol. abs/1702.00477, 2017.
  [Online]. Available: \url{http://arxiv.org/abs/1702.00477}
\BIBentrySTDinterwordspacing

\bibitem{DBLP:books/daglib/0023873}
\BIBentryALTinterwordspacing
M.~J{\"{u}}nger, T.~M. Liebling, D.~Naddef, G.~L. Nemhauser, W.~R. Pulleyblank,
  G.~Reinelt, G.~Rinaldi, and L.~A. Wolsey, Eds., \emph{50 Years of Integer
  Programming 1958-2008 - From the Early Years to the State-of-the-Art}.\hskip
  1em plus 0.5em minus 0.4em\relax Springer, 2010. [Online]. Available:
  \url{https://doi.org/10.1007/978-3-540-68279-0}
\BIBentrySTDinterwordspacing

\bibitem{Assign_2007}
D.~Pentico, ``Assignment problems: A golden anniversary survey,''
  \emph{European Journal of Operational Research}, vol. 176, pp. 774--793, 01
  2007.

\bibitem{Yuan2018}
Y.~Yuan, Y.~S. Ong, A.~Gupta, and H.~Xu, ``{Objective Reduction in
  Many-Objective Optimization: Evolutionary Multiobjective Approaches and
  Comprehensive Analysis},'' \emph{IEEE Transactions on Evolutionary
  Computation}, vol.~22, no.~2, pp. 189--210, 2018.

\bibitem{cheng2018benchmark}
R.~Cheng, M.~Li, Y.~Tian, X.~Xiang, X.~Zhang, S.~Yang, Y.~Jin, and X.~Yao,
  ``Benchmark functions for the cec'2018 competition on many-objective
  optimization,'' Tech. Rep., 2018.

\bibitem{Distance_Matrix_2014}
M.~Al-Neama, N.~Reda, and F.~Ghaleb, ``An improved distance matrix computation
  algorithm for multicore clusters,'' \emph{BioMed research international},
  vol. 2014, p. 406178, 06 2014.

\bibitem{Zitzler04indicator-basedselection}
E.~Zitzler and S.~Künzli, ``Indicator-based selection in multiobjective
  search,'' in \emph{in Proc. 8th International Conference on Parallel Problem
  Solving from Nature (PPSN VIII}.\hskip 1em plus 0.5em minus 0.4em\relax
  Springer, 2004, pp. 832--842.

\bibitem{Zitzler01spea2:improving}
E.~Zitzler, M.~Laumanns, and L.~Thiele, ``Spea2: Improving the strength pareto
  evolutionary algorithm,'' Tech. Rep., 2001.

\bibitem{Brockhoff:2019:Platypus}
\BIBentryALTinterwordspacing
D.~Brockhoff and T.~Tu\v{s}ar, ``Benchmarking algorithms from the platypus
  framework on the biobjective bbob-biobj testbed,'' in \emph{Proceedings of
  the Genetic and Evolutionary Computation Conference Companion}, ser. GECCO
  '19.\hskip 1em plus 0.5em minus 0.4em\relax New York, NY, USA: ACM, 2019, pp.
  1905--1911. [Online]. Available:
  \url{http://doi.acm.org/10.1145/3319619.3326896}
\BIBentrySTDinterwordspacing

\bibitem{PyGMO}
D.~Izzo, ``Pygmo and pykep: Open source tools for massively parallel
  optimization in astrodynamics (the case of interplanetary trajectory
  optimization),'' 01 2012, pp.~--.

\bibitem{gurobi}
\BIBentryALTinterwordspacing
L.~Gurobi~Optimization, ``Gurobi optimizer reference manual,'' 2019. [Online].
  Available: \url{http://www.gurobi.com}
\BIBentrySTDinterwordspacing

\end{thebibliography}

\end{document}